\documentclass[conference]{IEEEtran}
\IEEEoverridecommandlockouts
\usepackage{cite}
\usepackage{amsmath,amssymb,amsfonts}
\usepackage{algorithmic}
\usepackage{graphicx}
\usepackage{textcomp}
\usepackage{xcolor}
\def\BibTeX{{\rm B\kern-.05em{\sc i\kern-.025em b}\kern-.08em
    T\kern-.1667em\lower.7ex\hbox{E}\kern-.125emX}}
\begin{document}

\title{Performance of object recognition\\ in wearable videos
\thanks{This research has been funded by FEDER/Ministerio de Ciencia, Innovación y Universidades - Agencia Estatal de Investigación/RTC-2017-6421-7.
}
}

\author{\IEEEauthorblockN{Alberto Sabater}
\IEEEauthorblockA{\textit{DIIS-I3A} \\
\textit{University of Zaragoza, Spain}\\
asabater@unizar.es}
\and
\IEEEauthorblockN{Luis Montesano}
\IEEEauthorblockA{\textit{Bitbrain Technologies} \\
\textit{University of Zaragoza, Spain}\\
luis.montesano@bitbrain.es}
\and
\IEEEauthorblockN{Ana C. Murillo}
\IEEEauthorblockA{\textit{DIIS-I3A} \\
\textit{University of Zaragoza, Spain}\\
acm@unizar.es}
}

\maketitle

\begin{abstract}
Wearable technologies are enabling plenty of new applications of computer vision, from life logging to health assistance. Many of them are required to recognize the elements of interest in the scene captured by the camera
This work studies the problem of object detection and localization on videos captured by this type of camera.
Wearable videos are a much more challenging scenario for object detection than standard images or even another type of videos, due to lower quality images (e.g. poor focus) or high clutter and occlusion common in wearable recordings. Existing work typically focuses on detecting the objects of focus or those being manipulated by the user wearing the camera. We  perform a more general evaluation of the task of object detection in this type of video, because numerous applications, such as marketing studies, also need detecting objects which are not in focus by the user. 
This work presents a thorough study of the well known YOLO architecture, that offers an excellent trade-off between accuracy and speed, for the particular case of object detection in wearable video. We focus our study on the public ADL Dataset, but we also use additional public data for complementary evaluations. 
We run an exhaustive set of experiments with different variations of the original architecture and its training strategy. Our experiments drive to several conclusions about the most promising directions for our goal and point us to further research steps to improve detection in wearable videos.

\end{abstract}



\section{Introduction}
\label{sec:intro}



Recent hardware developments have generalized the use of new affordable wearable devices that 
people can carry during any kind of activity. These devices (including smartphones, fitness trackers or personal cameras) generate a vast amount of information along with new necessities and opportunities to use them. What we do, what we like, where we go, our health status and much more is unnoticed information that could help us to get insights of our daily life. It has also applications in many domains like healthcare, marketing, personalized training or home automation. 
The most exploited task in prior work using this kind of data has been activity analysis \cite{schuldt2004recognizing, Efros:2003:RAD:946247.946720}. These works for example make use of multiple sensors \cite{Maurer, Ellis:2014:MPA:2638728.2641673}, or analyze the user hands and manipulated objects \cite{990977} to identify the user activity. 

Our work focuses on the task of 
object-detection, since it provides relevant information in addition to user activity estimation, for example in augmented reality or marketing applications. Moving away from the object of focus usually requires more general and robust object detection in wearable videos.
Working with this type of videos presents additional challenges to object detection in regular images, such as frame blurring, camera defocus, fast object movements or strong clutter and occlusion on objects to be recognized. 
Egocentric videos record scenes from a particular perspective (first person perspective) due to chest or head mounted cameras. This also causes object occlusion by the human body itself. Human-object interaction also makes objects change their look when being interacted with. For instance, a fridge or a microwave can be both opened or closed, full or empty, etc. See Fig. \ref{sec:intro}.

\begin{figure}[!tb]
    \centering
    \begin{tabular}{cc}
    \includegraphics[width=.47\linewidth]{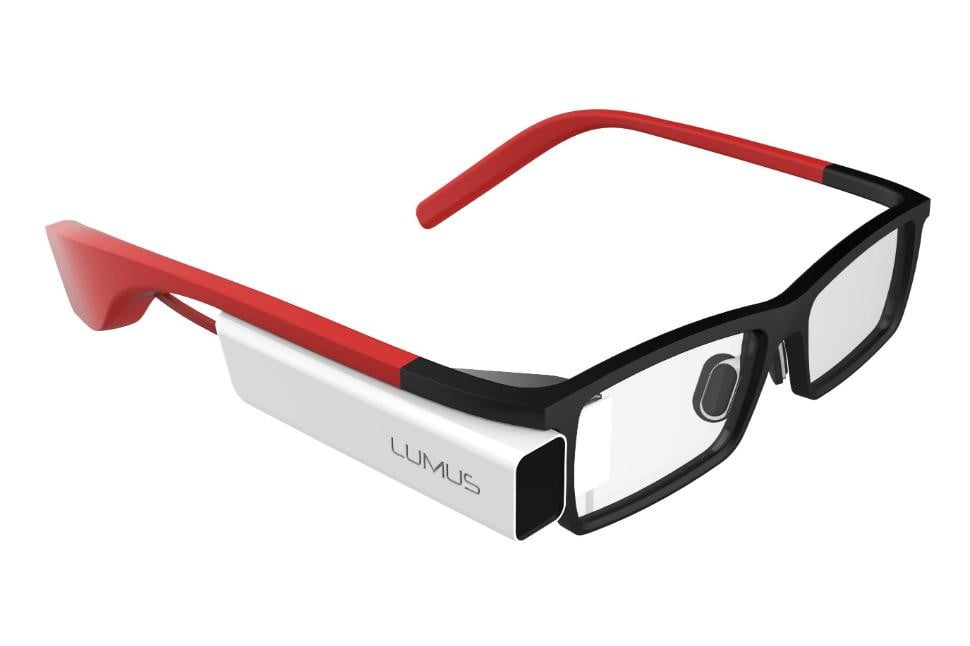} &
    \includegraphics[width=.47\linewidth]{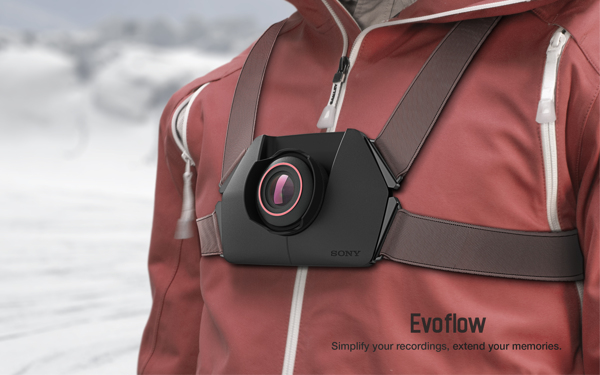}\\
    (a) & (b)\\
    \includegraphics[width=.47\linewidth]{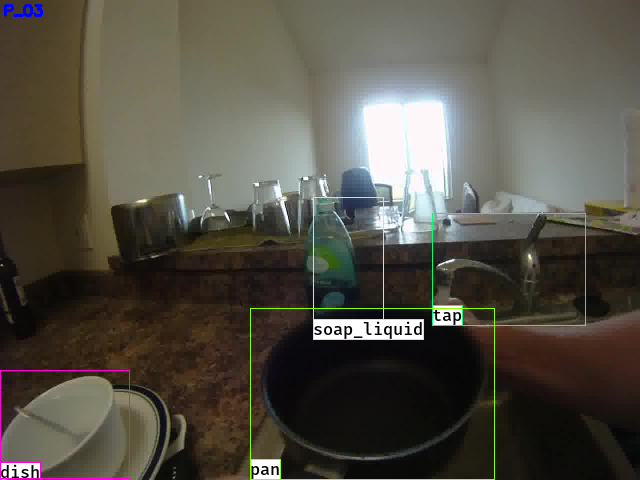} &
    \includegraphics[width=.47\linewidth]{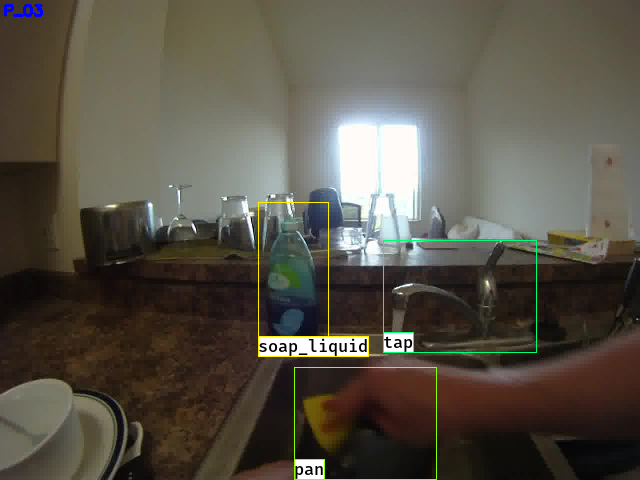}\\
    (c) & (d)\\
    \includegraphics[width=.47\linewidth]{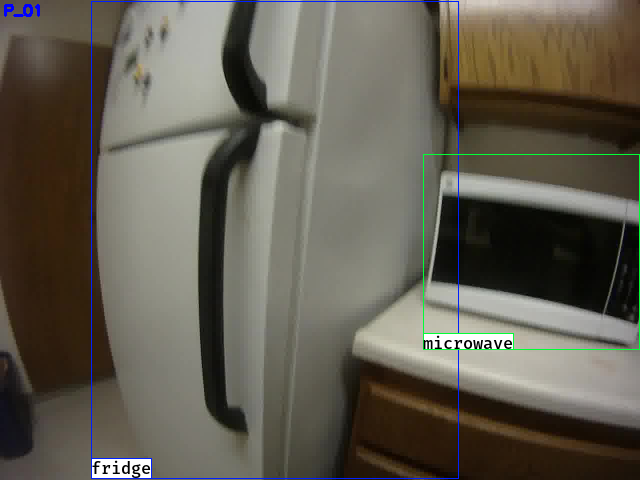} &
    \includegraphics[width=.47\linewidth]{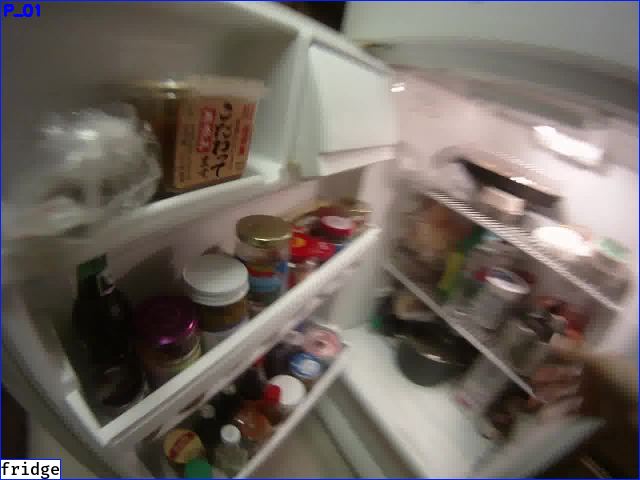}\\
    (e) & (f)\\
    \end{tabular}
\caption{Wearable video. (a), (b) wearable cameras. (c), (d) examples of typical strong object occlusions by the human body. (c), (d) human interaction produces strong appearance changes in the scene.}
    \label{fig:intro}
\end{figure}

Despite the impressive results in the field of object-recognition on conventional images and videos, object detection in egocentric videos recorded with wearable cameras are scarce. One of the reasons is the lack of properly labeled datasets.
This work studies the behaviour of  YOLO~\cite{DBLP:journals/corr/RedmonDGF15}  for object recognition in first-person perspective videos. YOLO is one of the most used object-detection architectures in real-time domains. We selected it for our study because it is one of the top performing methods for video object detection and presents a good trade-off between speed and accuracy.
We explore different models and data configurations and variations to measure YOLO's accuracy in different situations. We  discuss and propose additional ideas for further studies including available data and how to  build sub-sets that enable fine grained analysis of the results. 


\section{Related Work}
\label{sec:related}



\paragraph*{Object detection in images} 
State-of-the-art object-detectors on static images are commonly classified into two categories depending on the number of stages needed to predict the final bounding boxes. According to \cite{mod_conv_det}, single-stage solutions are faster than the ones that involve multiple steps, but they lack of the high accuracy of these second ones.

Multi-stage object-detectors follow a pipeline that first generate a set of Regions Of Interest (ROI) that will be classified and post-processed in posterior steps. Models from this family are mostly based in R-CNN \cite{DBLP:journals/corr/GirshickDDM13} and its posterior updates with Fast R-CNN\cite{DBLP:journals/corr/Girshick15}, Faster R-CNN\cite{DBLP:journals/corr/RenHG015} and R-FCN \cite{DBLP:journals/corr/DaiLHS16}.

In contrast to multi-stage detectors, single-pass detectors use a single Convolutional Neural Network trained end-to-end to predict object bounding boxes along with their labels in a single step. YOLO\cite{DBLP:journals/corr/RedmonDGF15}, SSD\cite{DBLP:journals/corr/LiuAESR15} and RetinaNet\cite{DBLP:journals/corr/abs-1708-02002} are the main single-stage detectors. Despite their lower accuracy, they offer faster predictions, depending on several factors such as the used feature extractor or the input image size that trade off speed and accuracy.

\paragraph*{Object detection in video}
Working with videos involves dealing with several additional issues with respect to detection in conventional images, like motion blur or camera defocus. This makes certain objects 
to be undetected or missclassified when we run a per-frame object-detection strategy. Since the release of the ImageNet \cite{ILSVRC15} object detection from video (VID) challenge, many approaches have been developed to handle these issues. 

Some of them involve post-processing methods like Seq-NMS\cite{DBLP:journals/corr/HanKPRBSLYH16} that uses high-scoring object detections from nearby frames to boost scores of weaker detections within the same clip or Seq-Bbox \cite{DBLP:journals/corr/HanKPRBSLYH16} that uses frame-level bounding box re-scoring to correct wrong detections and tubelet-level bounding box linking to infer boxes of missed detections.

Other approaches train end-to-end Neural Networks that perform feature aggregation such as  FGFA\cite{DBLP:journals/corr/ZhuWDYW17} that aggregate per-frame features along motion paths or D\&T\cite{Feichtenhofer17DetectTrack} that simultaneously perform detection and tracking.

Of particular relevance for our work are previous results in egocentric videos. The work in \cite{4408872} performs event classification by categorizing objects and classifying the environment of a frame. In \cite{Lee2015} important regions are recognized and used to do an egocentric video summarization. Finally,  \cite{Ramanan:2012:DAD:2354409.2355089} trains part-based models to detect and classify objects and uses temporal pyramid models to recognize actions.



\section{Object Recognition in Wearables}


In this work, we use YOLO v3\cite{DBLP:journals/corr/abs-1804-02767} to perform object detection in each frame of egocentric videos since it presents a good trade-off between speed and accuracy. 
Per-frame object detection is the first step to evaluate object recognition in this type of videos, the use of post-processing bounding boxes is left for next steps.
This section first summarizes the original architecture and then details all the training and execution variations we consider to evaluate the best performing options.








\subsection{YOLO}

YOLO v3 is the model from that improves YOLO9000\cite{DBLP:journals/corr/RedmonF16} to predict in one shot bounding box dimensions along with their object class. Thanks to its Fully Convolutional structure (without fully-connected layers), the neural net is able to take as input an image of any size as long as it is a multiple of 32 (due to its downsample/upsample pipeline). 

YOLO's architecture uses Darknet-53 as the backbone. This Network is made of residual blocks and shortcut connections and, trained on the ImageNet dataset\cite{ILSVRC15}, it achieves a similar performance as other architectures like ResNet-101 and ResNet-152\cite{DBLP:journals/corr/HeZRS15} with less computation power.

\begin{figure*}[!tb]
    \centering
    \includegraphics[width=.9\linewidth]{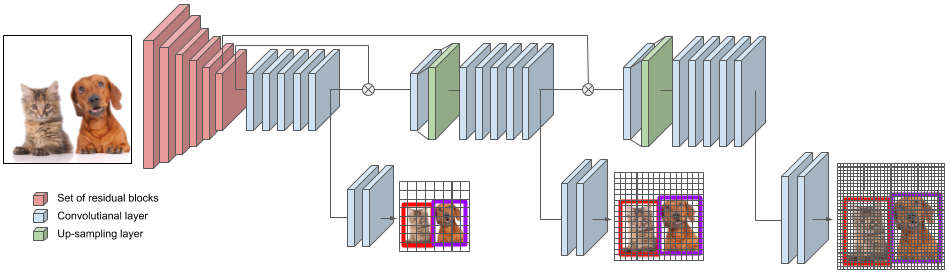}
    \caption{YOLO v3 architecture. 3 sets of bounding boxes are predicted for each input image. Deeper layers are focused on detecting smaller objects.}
    \label{fig:yolo_arch}
\end{figure*}

Inspired by Feature Pyramid Networks\cite{DBLP:journals/corr/LinDGHHB16}, YOLO makes predictions at three scales (see Fig. \ref{fig:yolo_arch}). To do so, it adds a set of convolutional layers after the backbone that outputs the first scale prediction. Next, it takes the feature maps from the 2 previous layers, upsamples it and merges it with feature maps from the backbone to be fed by another set of convolutional layers that outputs the next scale prediction. This process is repeated in a similar way to output the third scale prediction.

The final output predictions are 3D tensors with a shape of $N \times N \times [3 * (4 + 1 + C)]$. Each of these tensors divides the input image in a $N \times N$ grid, where each of its cells predicts $3$ boxes along with the probability of finding an object in that box and the probability of that object to belong to any of the $C$ classes contained in the given dataset.

Finally, the network predicts bounding boxes ($b_x$, $b_y$, $b_w$, $b_h$), using dimension clusters as anchor boxes, as follows:
\begin{eqnarray}
\label{eq:prediction}
\nonumber
&b_x = \sigma(t_x) + c_x \\ 
\nonumber
&b_y = \sigma(t_y) + c_y \\ 
\nonumber
&b_w = p_w e^{t_w} \\
&b_h = p_h e^{t_h},
\end{eqnarray}
\noindent where $t_x, t_y, t_w, t_h$ represent the predicted box, $c_x, c_y$ represent the offset from the top-left corner of the image and the bounding box prior has width and height $p_w, p_h$.
Once the bounding boxes of an input image have been predicted, those that are under a certain class probability threshold are filtered out. Then, the remaining ones are processed by non-max suppression to filter out overlapping bounding boxes.

To train this model, YOLO uses a loss function that can be split into three main components:
\begin{itemize}
    \item{Localization loss}: this score is composed by the error predicted both in the localization and size of bounding boxes. YOLO v3 measures it by using the sum of squared error loss but this varies in different implementations. Localization loss is set to 0 when no object has been predicted.
    \item{Confidence loss}: this is also split in two logistic functions that evaluate objects and background probabilities separately. This score should be 1 when a bounding box overlaps a ground truth object by more than any other bounding box prior.
    \item{Classification loss}: it uses independent logistic classifiers without softmax to perform multilabel classification. Then it can fit to datasets where one object can belong to different classes or when the labels are not consistent. This error is only measured when an object is detected.
\end{itemize}

\subsection{Model variations}

From this base implementation, we apply different modifications to evaluate its performance in different scenarios. Section \ref{sec:exp-results} provides more detailed information about the variations performed.

\paragraph*{Pretraining}
This is a common technique to take advantage of the patterns learned from other trainings and datasets. Using pretrained weights helps the model to achieve a better generalization to the specified domain and improves the convergence speed. In the present project we evaluate different pretrained weights and strategies to check the one that better suits our problem.

\paragraph*{Input size}
Due to the Fully Convolutional architecture of the Neural Network, the model's output size depends directly on the input image size. The bigger this is, the bigger output we get and more accurate information we can get. However, bigger input sizes involve higher prediction times, so it's important to find a good trade-off between these two elements.

\paragraph*{Alternative architectures}
Along with the main structure described before, the authors also provide the tiny-YOLO model~\cite{DBLP:journals/corr/RedmonDGF15}, a very small Neural Network focused on working on constrained environments, that makes predictions at only two scales and counts with a smaller backbone. 

\begin{figure}[!tb]
    \centering
    \includegraphics[width=\linewidth]{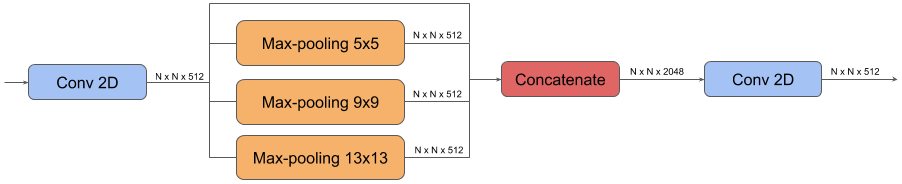}
    \caption{Architecture of a SPP block.}
    \label{fig:spp_block}
\end{figure}

Additionally, 
we have also used the base model extended by adding  Spatial Pyramid Pooling (SPP)\cite{DBLP:journals/pami/HeZR015}. SPP is a block (see Fig. \ref{fig:spp_block}) aimed to explode local multi-scale features to improve the final accuracy. It is made of a set of max-pooling layers that takes as input the set of feature maps generated by the backbone. Each of these layers pools its input, with stride 1, at different scales by using different window sizes. These three pooled feature maps are concatenated along with the original one to feed the next detection layers. The use of this SPP block helps the model to find objects at different scales.

\section{Experiments}
\label{sec:experiments}


% ...









\subsection {Datasets}

We have used the following three sets of data in our experiments because they are the ones that offer more and better annotated frames extracted from egocentric videos.


\paragraph*{ADL Dastaset \cite{Ramanan:2012:DAD:2354409.2355089}}
The main public dataset used in the experiments is the ADL Dastaset. It consists of 27000 frames extracted from 10 hours of video recorded with a chest-mounted GoPro of 20 people performing everyday activities in 20 different homes. These frames are densely annotated with activity and object labels. Due to the scope of this project, only the object annotations are used. These annotations consist of bounding boxes of 47 different objects.

\begin{figure}[!tb]
    \centering
    \begin{tabular}{cc}
    \includegraphics[width=.47\linewidth]{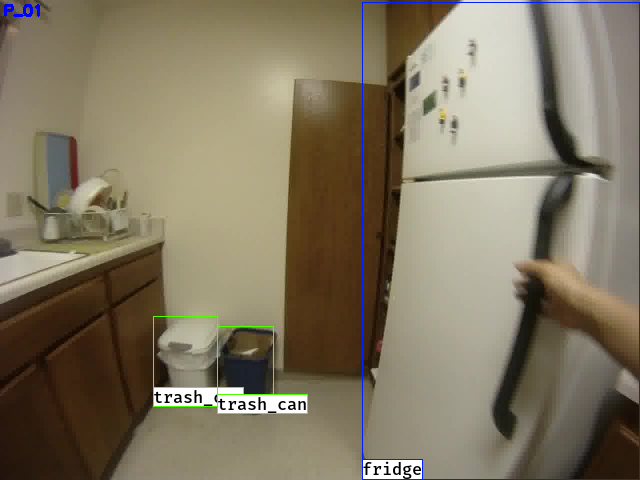} &
    \includegraphics[width=.47\linewidth]{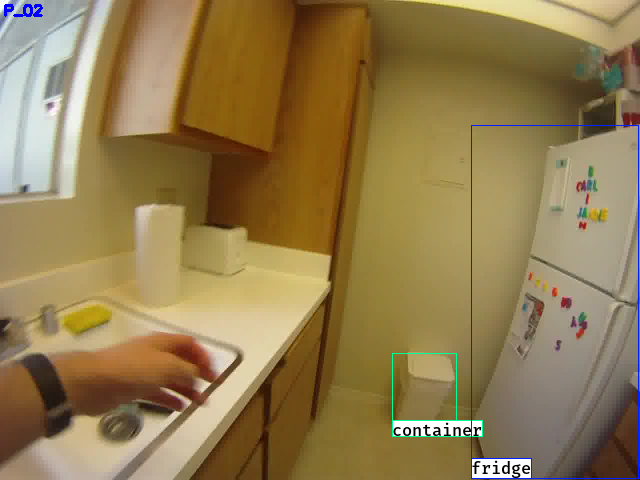}\\
    (a) & (b)\\
    \includegraphics[width=.47\linewidth]{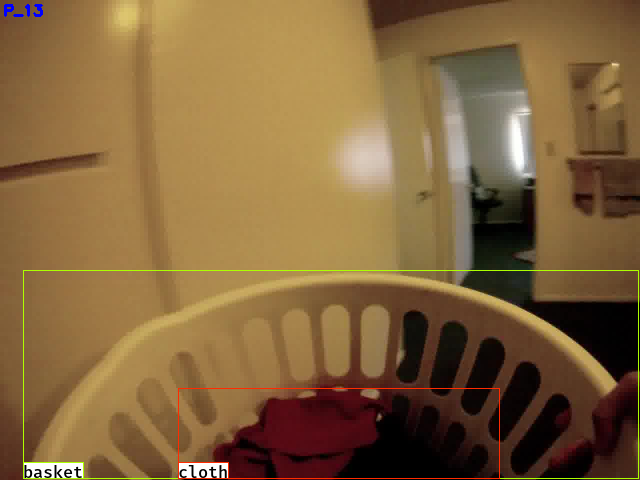} &
    \includegraphics[width=.47\linewidth]{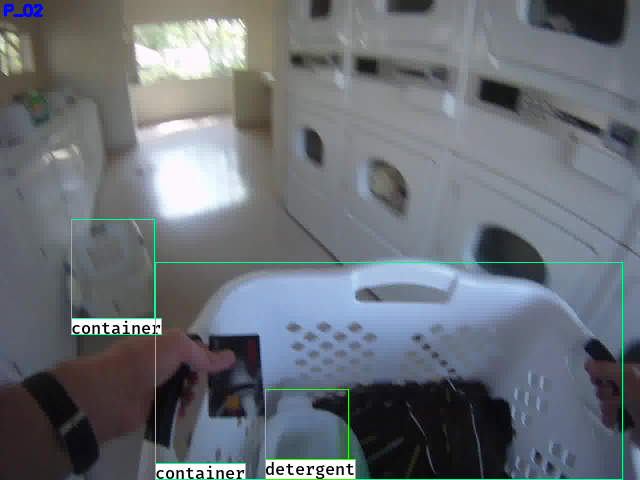}\\
    (c) & (d)\\
    \includegraphics[width=.47\linewidth]{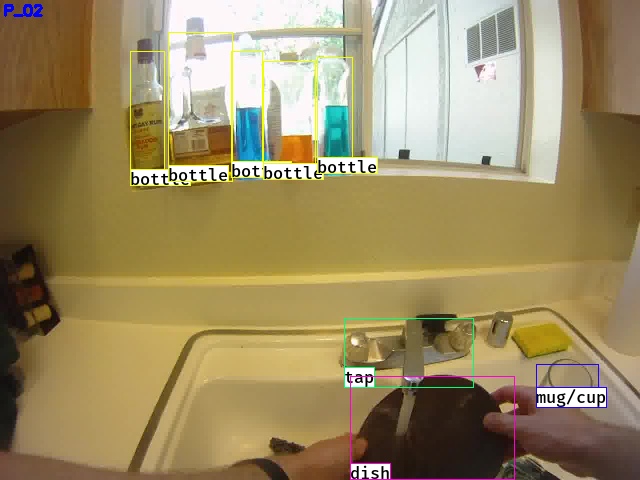} &
    \includegraphics[width=.47\linewidth]{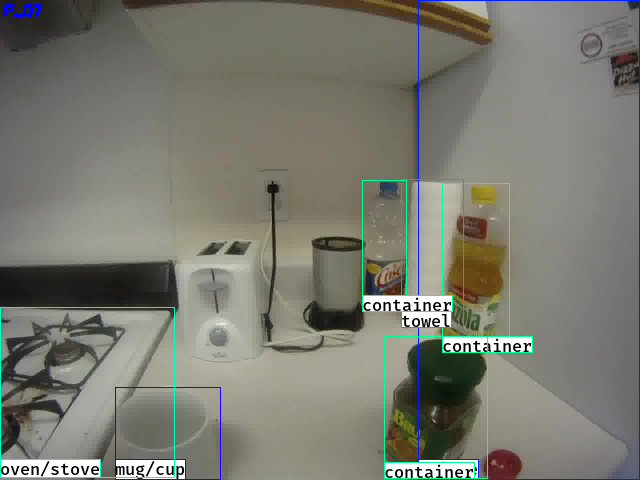}\\
    (e) & (f)\\
    \includegraphics[width=.47\linewidth]{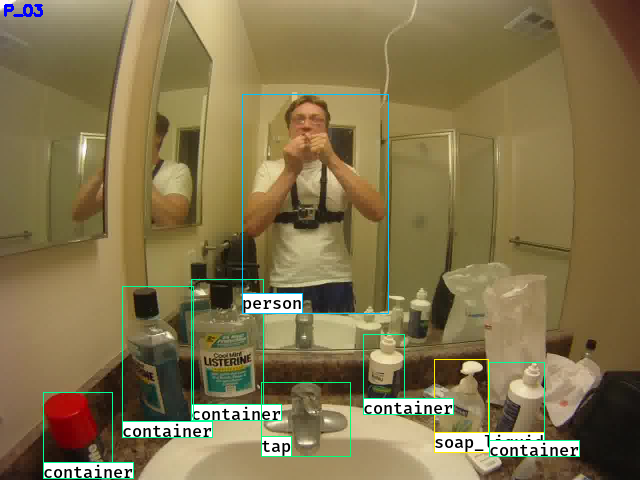} &
    \includegraphics[width=.47\linewidth]{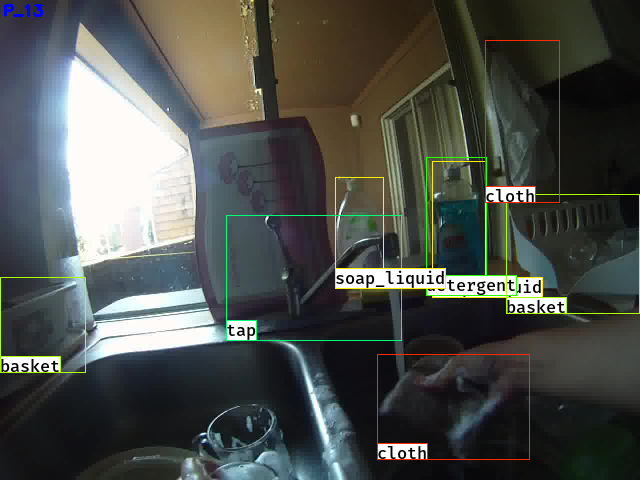}\\
    (g) & (h)\\
    \end{tabular}
\caption[Caption for LOF]{Samples of some labeling inconsistencies in different ADL Dataset videos. In images (a) and (b) the same object is labeled either as \textit{trash\_can} or \textit{container}. In (c) and (d) the same object is labeled either as a \textit{basket} or \textit{container}. In (e), (f), (g) and (h) a bottle can have the generic category \textit{bottle} or \textit{container}, or a specific one like \textit{soap\_liquid} or \textit{deterget}.}
    \label{fig:incons}
\end{figure}

Since the frames have been annotated by different people, there are some inconsistencies among certain videos, Fig. \ref{fig:incons} shows some examples of them. Each relevant object is not annotated in every frame where it occurs, and sometimes there are different class label annotations for the same object. For instance, classes like \textit{cell or cell\_phone}, \textit{shoe or shoes} and \textit{trash\_can or basket or container or large\_container} are used indistinctly.

To avoid having issues with this labeling inconsistency, we have build two sub-sets of the data with 27 (v2) and 8 (v3) classes respectively. Some classes have been merged or removed to obtain the new sub-sets in an attempt to have only consistent and unique object label  annotations. Table \ref{tab:dataset_classes} shows the whole set of class labels in each sub-set.

\begin{table}[!tb]
\centering
\begin{tabular}{|l|l|l|l|}
\hline
\multicolumn{4}{|c|}{{\bf Classes (27) from sub-set v2}} \\ \hline
\multicolumn{4}{|l|}{generic\_container (trash\_can, basket, container, large\_container)} \\ \hline
\multicolumn{2}{|l|}{bottle (perfume, bottle, milk/juice)} & book & cloth \\ \hline
\multicolumn{2}{|l|}{cell\_phone (cell, cell\_phone)} & dish & door \\ \hline
food/snack & fridge & kettle & laptop \\ \hline
knife/spoon/fork & microwave & mug/cup & oven/stove \\ \hline
\multicolumn{2}{|l|}{monitor/tv (monitor, tv)} & \multicolumn{2}{l|}{shoes (shoe, shoes)} \\ \hline
pan & person & pitcher & soap\_liquid \\ \hline
tap & tooth\_brush & tooth\_paste & towel \\ \hline
tv\_remote & washer/dryer &  &  \\ \hline
\multicolumn{4}{c}{}\\
\end{tabular}

\begin{tabular}{|l|l|l|}
\hline
\multicolumn{3}{|c|}{{\bf Classes (8) from sub-set v3}} \\ \hline
knife/spoon/fork            & laptop           & microwave    \\ \hline
\multicolumn{2}{|l|}{monitor/tv (monitor, tv)} & mug/cup      \\ \hline
pan                         & tap              & washer/dryer \\ \hline
\multicolumn{3}{c}{}\\
\end{tabular}

\caption{ Object classes in the built sub-sets of ADL dataset. Merged classes between brackets}
\label{tab:dataset_classes}

\end{table}

\paragraph*{EPIC-KITCHENS Dataset \cite{DBLP:journals/corr/abs-1804-02748}}
Additionally, the EPIC-KITCHENS Dataset has also been used as additional data in some of our tests. This dataset is composed by frames extracted from videos recorded in a head-mounted GoPro in 32 kitchens by 32 subjects. These frames are provided along with activity annotations and object bounding boxes. Even though this dataset is larger than the ADL one and shows a better annotation quality, the fact that only the \textit{active} objects (those that the subject is interacting with) are provided with annotations makes this dataset less useful for our evaluation purposes.

EPIC-KITCHENS contains 289 different classes, but a lot of them barely appear or are not included in the ADL dataset. So, 
we have also created a sub-set with the relevant classes, i.e., merged equivalent ones and removed classes not included in ADL. By doing so, two sub-sets have been built. EpicK$_{v1}$ contains  17 classes (show in \ref{tab:kitchen_classes}), and EpicK$_{v2}$ includes the same and additionally a \textit{food} class.
This food class has only been used in one of the sub-sets because it contains a set of 109 merged classes from the original dataset as well as the major part of the bounding boxes it contains. For this reason it is a matter of interest to check the model performance in both scenarios.

\begin{table}[!tb]
\centering
\begin{tabular}{|l|l|l|}
\hline
\multicolumn{3}{|c|}{{\bf Classes (17) from sub-set EpicK$_{v1}$}} \\ \hline
cutlery        & tap              & plate/bowl \\ \hline
fridge/freezer & salt/oil/vinegar & pan/pot    \\ \hline
bag/container  & cup/glass        & oven       \\ \hline
bottle/soup    & bin/basket       & cloth      \\ \hline
kettle         & coffee           & microwave  \\ \hline
colander       & washer           &            \\ \hline
\multicolumn{3}{c}{}\\
\end{tabular}
\caption{ Object classes in the built EpicK$_{v1}$ subset.}
\label{tab:kitchen_classes}
\end{table}

\paragraph*{Own unlabeled videos} we have also acquired additional recordings from a GoPro, independent of the previously mentioned datasets, to run additional validation experiments. In this data the user is looking at and interacting with objects included in the ADL Dataset. One video has been recorded with a chest-mounted camera (like the ADL data) and a second video has been recorded from a head-mounted camera (like the EPIC KITCHENS data).


\subsection{Experiments setup}

To evaluate the performance of our models we use the well known metric 
mean Average Precision (mAP) from PASCAL VOC \cite{Everingham:2010:PVO:1747084.1747104}, that considers a predicted bounding box as correct when its Intersection over Union (IoU) with the ground truth is above 0.5 ($mAP_{50}$). 

In contrast to the authors of the ADL dataset \cite{Ramanan:2012:DAD:2354409.2355089}, that train part-based models~\cite{Felzenszwalb:2010:ODD:1850486.1850574} using a leave-one-out cross-validation strategy, we train a Neural Network with the first 17 videos of the dataset, leaving the remaining three videos as a validation set. 
Therefore, the result in that work is a relevant baseline but it does not exactly show the same metric. Since their work already studied the variations across their dataset, we do not include that within our goals and opt to have a single fixed model to study the model variation influence.

Our \textit{base} solution is a YOLO v3 neural network built in TensorFlow (available at their authors website\footnote{https://pjreddie.com/darknet/yolo/}). We add different modifications to this base model to check its performance in different scenarios discussed in Section \ref{sec:exp-results}.

We start from the available base model pretrained with the weights learned on COCO dataset~\cite{DBLP:journals/corr/LinMBHPRDZ14} and fine-tuning it with the ADL data.
The pretrained layers are frozen during the first 15 epochs and both the input training and test images have a fixed size of 416x416.  All our experiments have been performed with a single NVIDIA Geforce RTX 2080 Ti GPU.\\

\subsection{Results}\label{sec:exp-results}
After an exhaustive study of the YOLO's performance in our different data sub-sets, we have found that the major update to our base solution is the use of multi-scale image training (discussed later). For a better generalization, we pretrain our Neural Network with COCO data,  and depending on the data sub-set, we have seen that using Spatial Pyramid Pooling is also beneficial.

Table \ref{tab:full} shows a summary of all the experiments we performed over our base solution described before. We vary the training data sub-set, the model architecture, its pretraining and the training input image size. The first row shows the performance of the base solution in the original ADL dataset (with no classes merged or removed). 
The relatively low mAP score (note the original set has 47 classes) is partially due to the high intra-class variability, i.e., since each video is recorded in a different home, the particular instances of the object classes can be very different.
We have also observed that the detections evaluated as incorrect are often produced by the label consistency issues discussed in previously. This is confirmed since the models trained with non-conflicting classes (v2 and v3), obtain much higher scores.

\begin{table}[!tb]
\centering
\begin{tabular}{|l|l|l|l||l|}
\hline
{\bf sub-set} & {\bf architecture} & {\bf pretraining} & {\bf input size} & {\bf mAP\textsubscript{50}} \\ \hline\hline
original & base & coco & 416 & 25.400 \\ \hline
original & spp & coco & 416 & 25.658 \\ \hline
original & base & coco & multi-scale & \textbf{26.298} \\ \hline
original & spp & coco & multi-scale & 26.219 \\ \hline\hline
v2 & base & coco & 416 & 38.794 \\ \hline
v2 & base & -- & 416 & 25.948 \\ \hline
v2 & base & backbone & 416 & 31.441 \\ \hline
v2 & base & epicK$_{v1}$ & 416 & 31,449 \\ \hline
v2 & base & epicK$_{v2}$ & 416 & 30,934 \\ \hline
v2 & base & coco & 320 & 36.771 \\ \hline
v2 & base & coco & 608 & 39.188 \\ \hline
v2 & spp & coco & 320 & 36.417 \\ \hline
v2 & spp & coco & 416 & 39.496 \\ \hline
v2 & spp & coco & 608 & 38.850 \\ \hline
v2 & base & coco & multi-scale & 38.946 \\ \hline
v2 & spp & coco & multi-scale & \textbf{39.694} \\ \hline
v2 & tiny & coco & 416 & 26.989 \\ \hline\hline
v3 & base & coco & 416 & 51.145 \\ \hline
v3 & base & -- & 416 & 39.803 \\ \hline
v3 & base & backbone & 416 & 46.773 \\ \hline
v3 & base & epicK$_{v1}$ & 416 & 46,362 \\ \hline
v3 & base & epicK$_{v2}$ & 416 & 46,437 \\ \hline
v3 & base & coco & 320 & 48.326 \\ \hline
v3 & base & coco & 608 & 51.692 \\ \hline
v3 & spp & coco & 320 & 48.304 \\ \hline
v3 & spp & coco & 416 & 51.514 \\ \hline
v3 & spp & coco & 608 & 50.705 \\ \hline
v3 & base & coco & multi-scale & \textbf{51.839} \\ \hline
v3 & spp & coco & multi-scale & 51,570 \\ \hline
v3 & tiny & coco & 416 & 42.798 \\ \hline

\multicolumn{5}{c}{}\\
\end{tabular}
\caption{Configuration of all YOLO variations evaluated and the corresponding $mAP_{50}$ metric. 
}
\label{tab:full}
\end{table}

Analyzing the results in more detail, we observe a huge difference in the object-detection accuracy between different object classes. 
We 
summarize this in Table \ref{tab:raw}, that compares the most relevant results from our experiments with the results presented as baseline by the ADL dataset authors.
As we can see, our base choice with the updates discussed 
outperform the dataset authors' solution based on part-based models.

\begin{table}[!tb]
\centering
\begin{tabular}{@{}|l||l||p{0.9cm}|p{0.9cm}@{}|p{1.3cm}@{}|p{1.3cm}@{}|}
\hline
{\bf Object} & \multicolumn{5}{c|}{\bf Architecture used for the detection}\\
\cline{2-6}
{\bf class} & *ADL~\cite{Ramanan:2012:DAD:2354409.2355089} & Base YOLO & SPP YOLO & Base + multi-scale & SPP + multi-scale \\ \hline
tap             & 40.4 $\pm$ 24.3	& 80.85	& 80.46	& 78.67	& \textbf{81.18} \\ \hline
soap\_liquid     & 32.5 $\pm$ 28.8	& 53.20	& 43.36	& 54.16	& \textbf{60.08} \\ \hline
fridge          & 19.9 $\pm$ 12.6	& 71.75	& 69.23	& \textbf{73.40}	& 68.96 \\ \hline
microwave       & 43.1 $\pm$ 14.1	& 74.52	& 73.84	& \textbf{78.07}	& 70.24 \\ \hline
oven/stove      & 38.7 $\pm$ 22.3	& 61.87	& 58.55	& \textbf{63.07}	& 61.19 \\ \hline
bottle          & 21.0 $\pm$ 27.0	& 17.38	& \textbf{25.82}	& 25.68	& 19.61 \\ \hline
kettle          & 21.6 $\pm$ 24.2	& 24.17	& \textbf{35.31}	& 30.28	& 29.53 \\ \hline
mug/cup         & 23.5 $\pm$ 14.8	& 22.50	& 16.49	& \textbf{24.89}	& 20.88 \\ \hline
washer/dryer    & 47.6 $\pm$ 15.7	& 63.11	& 69.11	& \textbf{69.67}	& 67.70 \\ \hline
tv              & 69.0 $\pm$ 21.7	& 81.39	& 80.40	& \textbf{84.20}	& 82.61 \\ \hline
\multicolumn{6}{p{8.5cm}}{*The results presented in that work are an average of several train/test splits, while our results use a unique train/test split. They also perform object-recognition only for 24 classes while our results come from a training with all categories.}\\
\multicolumn{6}{c}{}\\
\end{tabular}
\caption{
Comparison of object-detection results between the ADL author's model, our base model ($Base$) and different studied variations (Spatial Pyramid Pooling and multi-scale training). All trained with the original ADL Dataset.
}
\label{tab:raw}
\end{table}

In the remainder of this section, we discuss in more detail our exhaustive experimentation and variations of the base model, where we use only the proposed ADL sub-sets to mitigate the effect of inconsistencies in the labeling.


\paragraph*{Influence of different pretraining strategies} 
To improve the model generalization we have also tested different fine-tuning strategies. This set of tests involve training the NN with random weights initialization, pretraining just the backbone with weights obtained from an ImageNet training or use the weights learned from our EPIC-KITCHENS sub-set versions. Table~\ref{tab:full} shows (first five rows of v2 and v3) that these modifications result in a decrease of 7 to 13 and 5 to 11 of the mAP score when training with the sub-sets v2 and v3 respectively. 
Therefore, it is clear that pretraining the full network with COCO weights brings significant benefits. \\

\begin{figure}[!tb]
    \centering
    \begin{tabular}{@{}c@{\hspace{2mm}}c}
    \includegraphics[width=.49\linewidth]{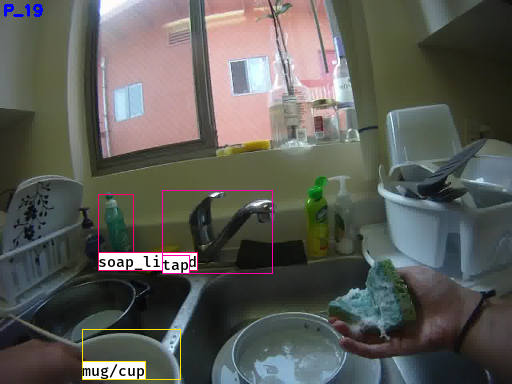} &
    \includegraphics[width=.49\linewidth]{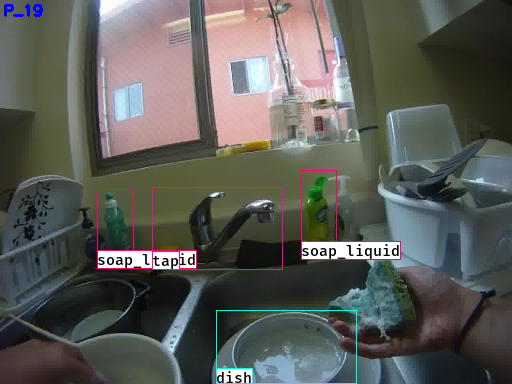}\\
    (a) & (b)\\
    \multicolumn{2}{c}{\includegraphics[width=.49\linewidth]{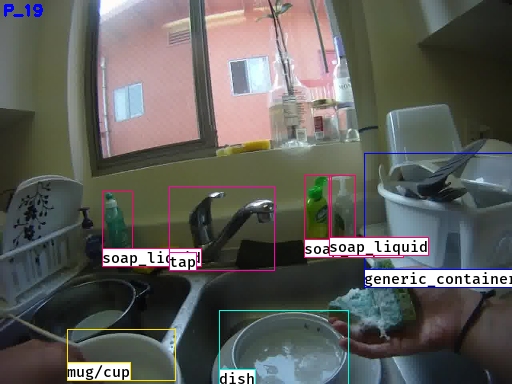}}\\
    \multicolumn{2}{c}{(c)}\\
    \end{tabular}
\caption{
Detection results on the same frame when training and testing with different image sizes: a) 320x320, b) 416x416, c) 608x608.}
    \label{fig:img_size}
\end{figure}

\paragraph*{Analysis of the training input image size}
We have trained YOLO with different fixed input sizes, and made the predictions with that same size. 
Fig. \ref{fig:img_size} shows 
an illustrative example of the difference in accuracy among them. As it could be expected, high resolution images provide more information for the object localization and classification. Additionally, larger inputs facilitate to detect objects that are too close to each other or too small but requires higher computational resources. Results in Table \ref{tab:full} (see the mAP in the rows 1, 6 and 7 for the v2 and v3 data sub-sets in Table \ref{tab:full}) shows that our base input image size (416x416) achieves a score much higher that the smaller input and only slightly worst than the bigger one. Therefore, the base input image size provides a good trade-off between accuracy and speed.\\

\paragraph*{Analysis of different YOLO architectures}\label{sec:exp_spp}
Here, we discuss the results for the different neural network architectures. Besides the base model with 75 Convolutional layers, we have also tested the \textit{tiny} architecture (tiny-yolo) proposed by the authors, with 13 Convolutional layers and a model that includes a Spatial Pyramid Pooling \cite{DBLP:journals/pami/HeZR015} after the backbone (SPP-YOLO) and is composed by 76 Convolutional layers. Tiny-yolo is able to get much faster results than standard YOLO but at the cost of lower 
scores. SPP-YOLO is able to improve the base architecture results without a significant increment in the prediction speed. See Table \ref{tab:speed-acc}.\\

\paragraph*{SPP-YOLO performance with different input sizes} To get a deeper insight about the SPP performance, we have compared this architecture with the base model and different input image sizes. SPP-YOLO is able to detect objects at different scales due to the use of feature maps processed by several pooling layers with different configurations. Table \ref{tab:full} shows how this update improves the base results achieving a mAP score with the base input image size competitive or better to the one achieved with a larger input. That means, better performance with less computing cost. However, the SPP results obtained with different input shapes do not show that improvement. That could be solved by tuning the maxpooling configurations in the SPP hyperparameters. Higher pooling windows could fit better larger images, and lower pooling windows could fit better smaller image sizes, but that is left for a future research.\\

\paragraph*{Analysis of multi-scale training}
Thanks to the Fully Convolutional YOLO's architecture, it can be trained with batches of images with different sizes. By feeding the model with different image sizes, it learns features at different scales, making generalization to different object sizes easier. In this way we have performed a set of tests that involves training with random variable image sizes, but the evaluation is performed with a fixed shape of 416x416. 

As a result, in Table \ref{tab:full} we observe a significant improvement of the performance of multi-scale trainings compared with the base model in each of the sub-sets, that also outperforms the SPP results. We also observe how multi-scale training with the SPP architecture achieves comparable or better results than the ones obtained with the base architecture.\\

\paragraph*{Analysis of the accuracy/speed trade-off}
Table \ref{tab:speed-acc} shows the effect of model complexity and input image size on the model precision. 

The input image size is a key factor to get fast predictions. In this way, we can see how an input size of 608x608 obtains the best results in general terms but it doubles the FLOPS needed of the smaller input size. In addition, the accuracy obtained with an input size of 416x416 is competitive with it and is a good compromise in this case.

Tiny-yolo is a really small model that is able to perform fast predictions but its accuracy is far from the base model results. This architecture could be able to fit systems with low specifications where efficiency is more important than accuracy.

Finally, we can check how the SPP architecture outperforms the base model with only increasing the Float Operations by 0.5\%.\\

\begin{table*}[!tb]
\centering
\begin{tabular}{|l|l|r|r|l|l|}
\hline
Architecture & Input size & \multicolumn{1}{l|}{FLOPS*} & \multicolumn{1}{l|}{Params} & mAP\textsubscript{50} v2 & mAP\textsubscript{50} v3 \\ \hline
base & 320 x 320 & 39.06 Bn & 61.72 M & 36.771 & 48.326 \\ \hline
base & 416 x 416 & 65.80 Bn & 61.72 M & 38.794 & 51.145 \\ \hline
base & 608 x 608 & 140.21 Bn & 61.72 M & 39.188 & 51.692 \\ \hline
SPP & 320 x 320 & 39.29 Bn & 62.77 M & 36.417 & 48.304 \\ \hline
SPP & 416 x 416 & 66.19 Bn & 62.77 M & \textbf{39.496} & \textbf{51.514} \\ \hline
SPP & 608 x 608 & 141.02 Bn & 62.77 M & 38.85 & 50.705 \\ \hline
tiny & 416 x 416 & 5.53 Bn & 8.74 M & 26.989 & 42.798 \\ \hline
\multicolumn{6}{c}{*FLOPS have been calculated with sub-set v2 but the numbers barely change in the sub-set v3.}\\
\multicolumn{6}{c}{}\\
\end{tabular}
\caption[Caption for LOF]{
Trade-off 
analysis between
speed and accuracy for object-detection with models trained on v2 and v3 sub-sets.}
\label{tab:speed-acc}
\end{table*}

\paragraph*{Qualitative analysis in additional scenarios}

\begin{figure}[!tb]
    \centering
    \begin{tabular}{cc}
    \includegraphics[width=.45\linewidth]{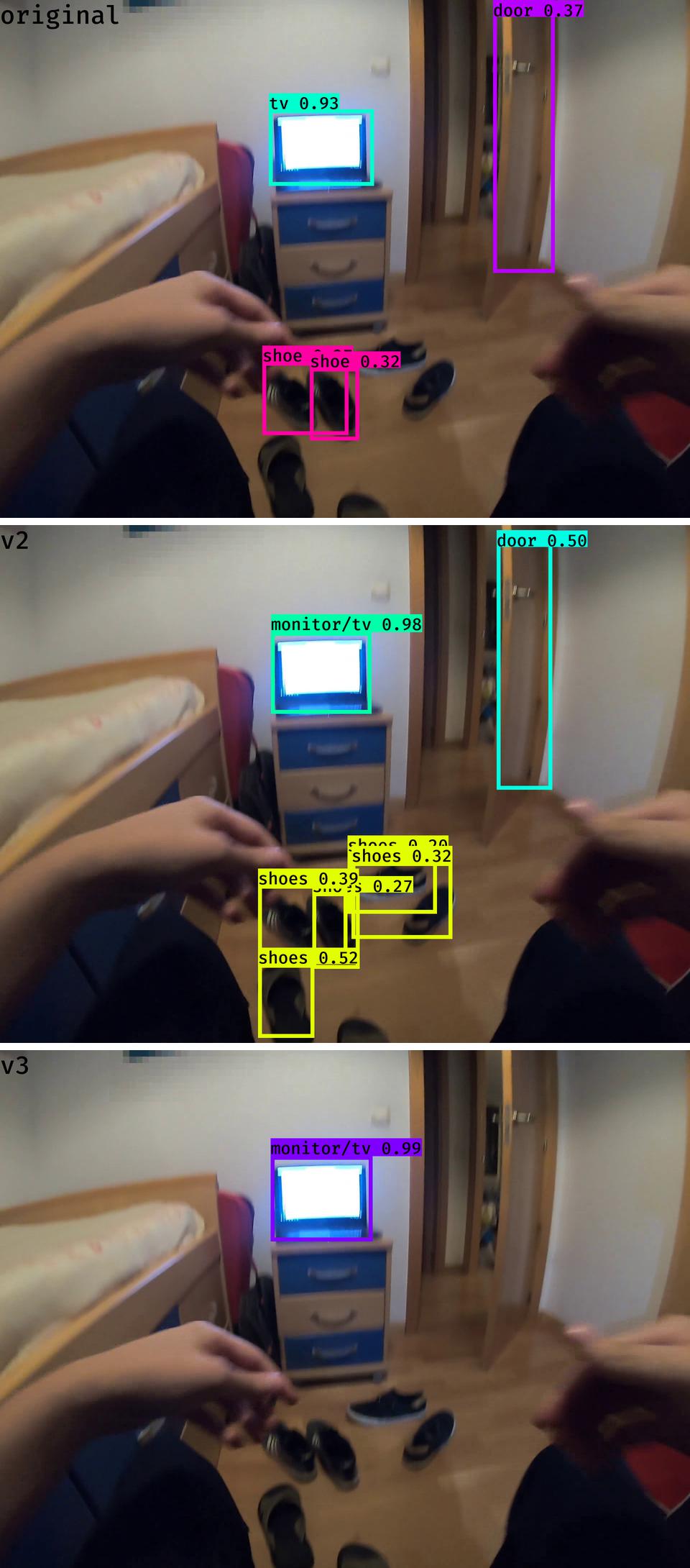} &
    \includegraphics[width=.45\linewidth]{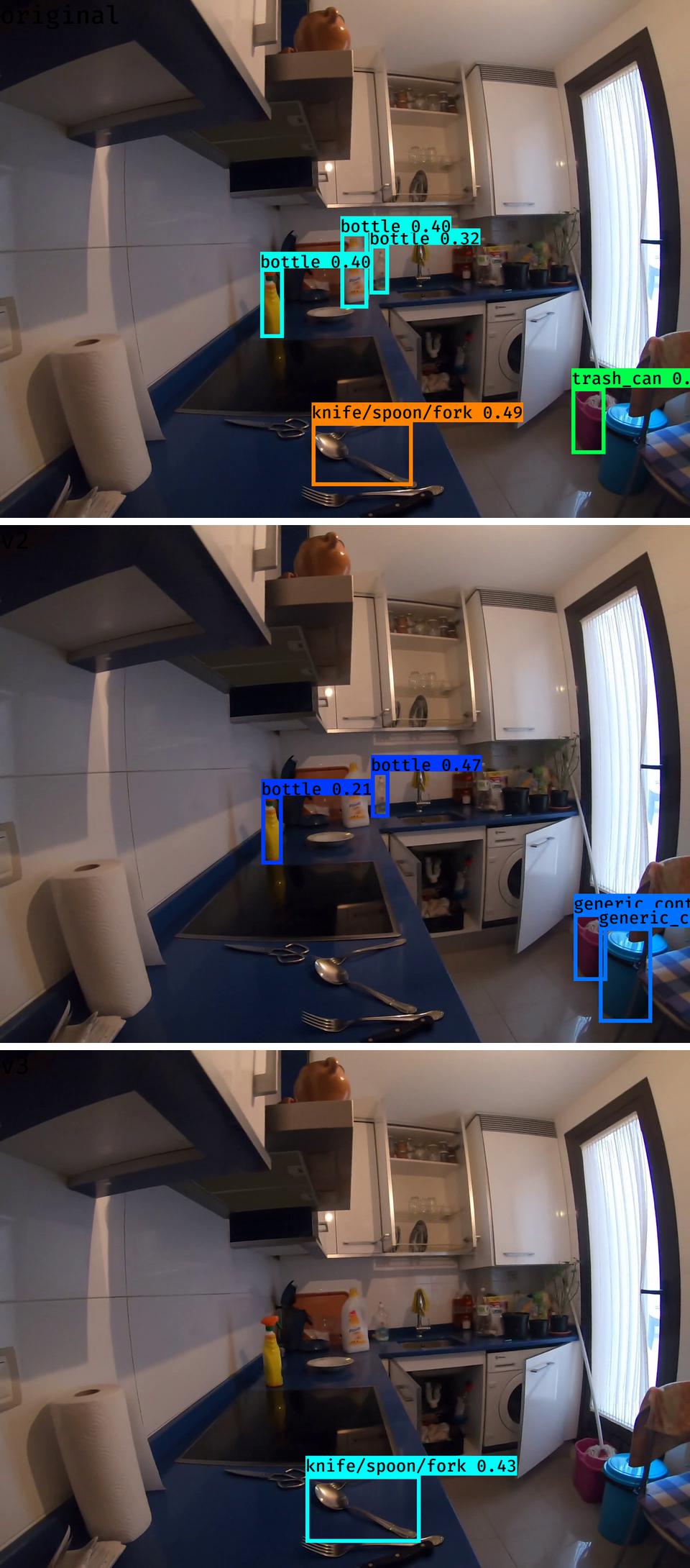}\\
    (a) & (b)\\
    \includegraphics[width=.45\linewidth]{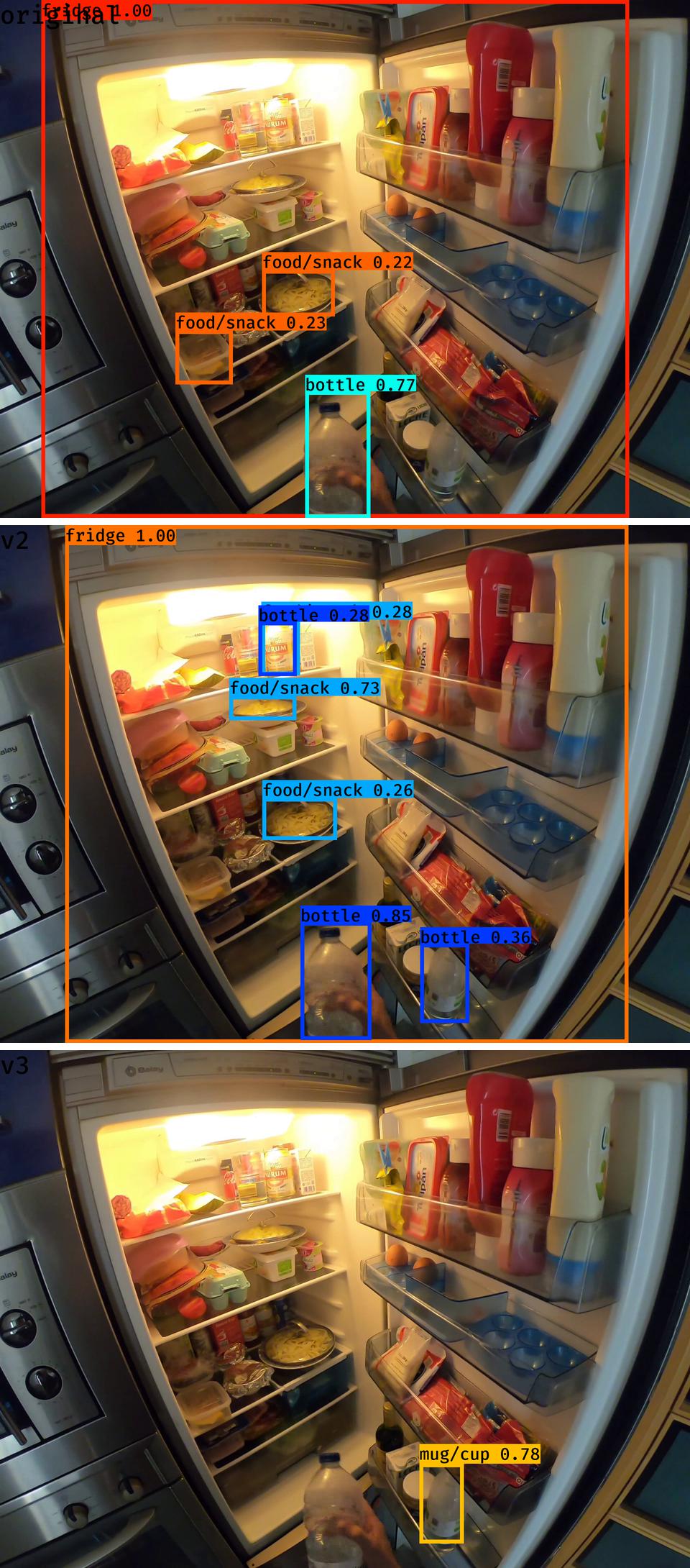} &
    \includegraphics[width=.45\linewidth]{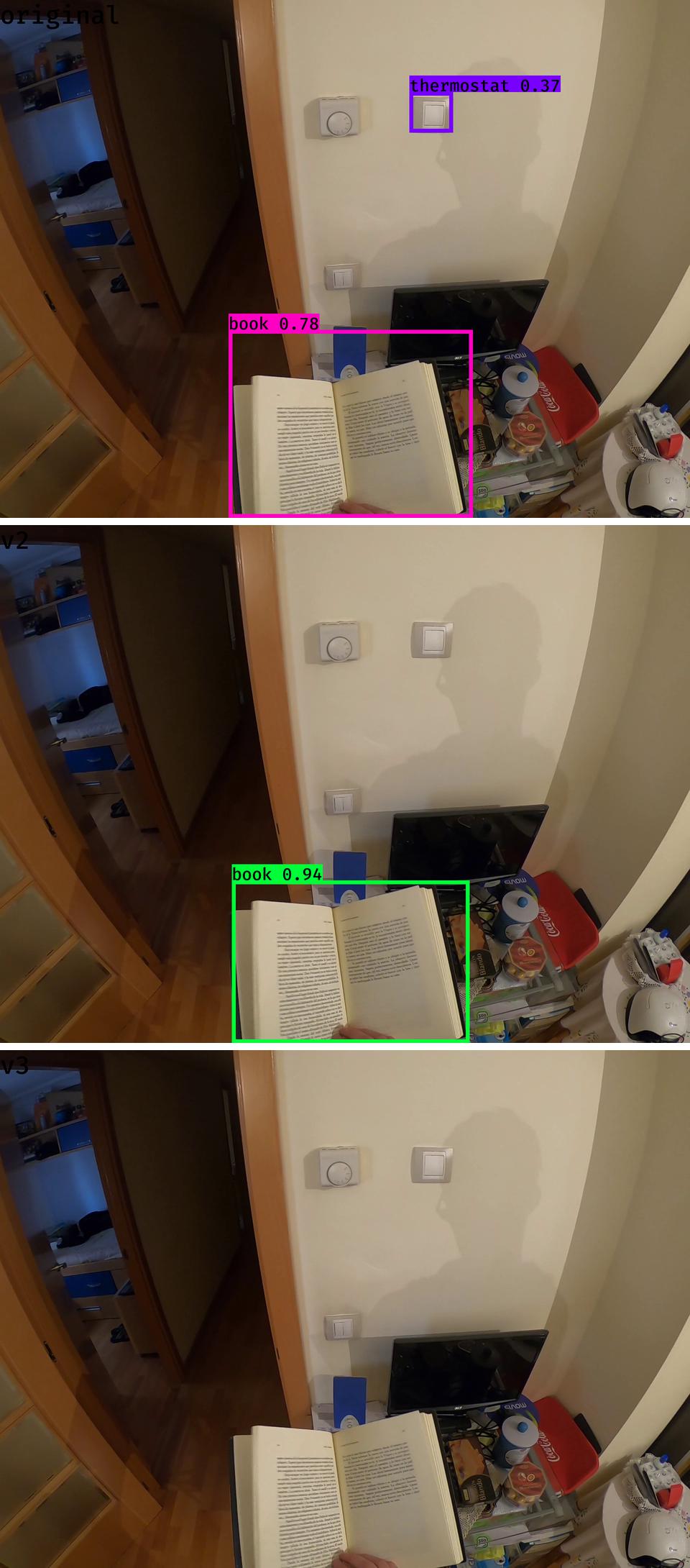}\\
    (c) & (d)\\
    \end{tabular}
\caption{Sample frames from additional unlabeled videos.
(a), (b) correspond to chest-mounted camera video. (c), (d) correspond to the head-mounted camera video. First, second and third rows correspond to predictions of the original ADL dataset classes, and the v2 and v3 subsets respectively. }
    \label{fig:home_test}
\end{figure}

We have validated our best models trained with each of our custom data sub-sets, as well as the one trained with the original ADL Dataset in our unlabeled videos. Fig.~\ref{fig:home_test} shows some sample predictions with the objects found along those videos.
Additional results on these videos, along with the code to replicate this work experiments are available online~\footnote{https://sites.google.com/a/unizar.es/filovi}.

Even though the style of this new environment slightly differs from the ones in the ADL Dataset, we observe a pretty good object generalization. However, since our models have been trained only with frames extracted from videos recorded with chest-mounted cameras they find hard to generalize to the perspective obtained with the head-mounted camera. This issue increases the chances of object misdetection or misclassification.

In general terms, we observe how the model trained with the original dataset is able to generate many more bounding boxes that the ones trained with other data sub-sets, but it does not perform as well as the other models at the actual object classification. This issue appears when detecting specific unusual classes like \textit{vacuum} or \textit{bed} and it gets more false positives with the ones that have label inconsistencies.

The model trained with the sub-set v3 
shows less accurate object-localization  but better object-classification that training with the original dataset. It also generates more robust and uniform bounding boxes and class labels over time. Models trained with other data sub-sets perform worse at generating uniform results over the video sequence (i.e. contiguous frames switch classes more often).

Finally, the model trained with the sub-set v3 generates more robust and uniform bounding boxes over time, along with their classes. However, it performs worse at detecting objects since it does not predicts as many boxes as expected. This issue could happen because of the existence of less bounding boxes during training, so the network focuses more on the background rather than in the objects. We believe that this could be fixed by tuning the loss hyper-parameters.



\section{Conclusion}

This work shows a detailed evaluation of the YOLO architecture and its performance for object-detection in wearable videos (per-frame). We have discussed the main issues in this kind of videos and how we deal with label inconsistencies in existing relevant datasets.

We have performed several modifications to the model architecture and training strategy of our YOLO base experiment.  
After our exhaustive validation, we have found that the pretraining strategy, with COCO weights in this case, is a key element for a good model performance and fast convergence. We have also shown how to improve the object-scale generalization by using Spatial Pyramid Pooling and multi-scale training, without increasing  training or prediction speed. 
Besides, we have evaluated the performance of the YOLO models in a new scenario and discussed the benefits and drawbacks of each of them and how they handle different view points.


As work-in-progress, there are additional ideas that would help YOLO to take advantage of temporal patterns in the frames. In particular, performing bounding boxes post-processing and feature aggregation. Another open research line is to modify the  YOLO loss function  to better fit the nature of the wearable data. For instance, the different weights of the loss components could be adjusted to balance the coordinates, background or object relevance during training.


\bibliographystyle{IEEEtran}
\bibliography{biblio}

\end{document}